\documentclass[11pt,a4paper]{article}
\usepackage[hyperref]{acl2018}
\usepackage{times}
\usepackage{latexsym}
\usepackage{mathrsfs}
\usepackage{amssymb}
\usepackage{graphicx}
\usepackage{adjustbox} 
\usepackage{booktabs} 
\usepackage{multirow} 

\usepackage{amsfonts}
\usepackage{amsmath}

\usepackage{amssymb}

\usepackage{url}

\usepackage{float} 

\newtheorem{definition}{Definition}

\aclfinalcopy 
\def\aclpaperid{1549} 

\title{Do Neural Network Cross-Modal Mappings Really Bridge Modalities?}

\author{Guillem Collell \\
  Department of Computer Science \\
  KU Leuven \\
  {\tt gcollell@kuleuven.be} \\\And
  Marie-Francine Moens \\
  Department of Computer Science \\
  KU Leuven \\
  {\tt sien.moens@cs.kuleuven.be} \\}

\date{}

\begin{document}
\maketitle
\begin{abstract}
Feed-forward networks are widely used in cross-modal applications to bridge modalities by mapping distributed vectors of one modality to the other, or to a shared space. The predicted vectors are then used to perform e.g., retrieval or labeling. Thus, the success of the whole system relies on the ability of the mapping to make the neighborhood structure (i.e., the pairwise similarities) of the predicted vectors akin to that of the target vectors. However, whether this is achieved has not been investigated yet. Here, we propose a new similarity measure and two ad hoc experiments to shed light on this issue. In three cross-modal benchmarks we learn a large number of language-to-vision and vision-to-language neural network mappings (up to five layers) using a rich diversity of image and text features and loss functions. Our results reveal that, surprisingly, the neighborhood structure of the predicted vectors consistently resembles more that of the input vectors than that of the target vectors. In a second experiment, we further show that untrained nets do not significantly disrupt the neighborhood (i.e., semantic) structure of the input vectors.
\end{abstract}

\section{Introduction}

Neural network mappings are widely used to bridge modalities or spaces in cross-modal retrieval \cite{qiao2017visually,wang2016comprehensive,zhang2016fast}, zero-shot learning \cite{lazaridou2015combining,lazaridou2014wampimuk,socher2013zero} in building multimodal representations \cite{collell2017imagined} or in word translation \cite{lazaridou2015hubness}, to name a few. Typically, a neural network is firstly trained to predict the distributed vectors of one modality (or space) from the other. 
At test time, some operation such as retrieval or labeling is performed based on the nearest neighbors of the predicted (mapped) vectors. For instance, in zero-shot image classification, image features are mapped to the text space and the label of the nearest neighbor word is assigned. Thus, the success of such systems relies entirely on the ability of the map to make the predicted vectors similar to the target vectors in terms of semantic or neighborhood structure.\footnote{We indistinctly use the terms \textit{semantic structure}, \textit{neighborhood structure} and \textit{similarity structure}. They refer to all pairwise similarities of a set of $N$ vectors, for some similarity measure (e.g., Euclidean or cosine).}
However, whether neural nets achieve this goal in general has not been investigated yet. In fact, recent work evidences that considerable information about the input modality propagates into the predicted modality \cite{collell2017imagined,lazaridou2015combining,frome2013devise}.

To shed light on these questions, we first introduce the (to the best of our knowledge) first existing measure to quantify \textit{similarity between the neighborhood structures of two sets of vectors}. Second, we perform extensive experiments in three benchmarks where we learn image-to-text and text-to-image neural net mappings using a rich variety of state-of-the-art text and image features and loss functions. Our results reveal that, contrary to expectation, the semantic structure of the mapped vectors consistently resembles more that of the input vectors than that of the target vectors of interest. In a second experiment, by using six concept similarity tasks we show that the semantic structure of the input vectors is preserved after mapping them with an untrained network, further evidencing that feed-forward nets naturally preserve semantic information about the input. Overall, we uncover and rise awareness of a largely ignored phenomenon relevant to a wide range of cross-modal / cross-space applications such as retrieval, zero-shot learning or image annotation.

Ultimately, this paper aims at: (1) Encouraging the development of better architectures to bridge modalities / spaces;
(2) Advocating for the use of semantic-based criteria to evaluate the quality of predicted vectors  
such as the neighborhood-based measure proposed here, instead of purely geometric measures such as mean squared error (MSE).

\section{Related Work and Motivation}
\label{related_work}

Neural network and linear mappings are popular tools to bridge modalities in \textit{cross-modal retrieval} systems. \citet{lazaridou2015combining} leverage a text-to-image linear mapping to retrieve images given text queries. \citet{weston2011wsabie} map label and image features into a shared space with a linear mapping to perform \textit{image annotation}. Alternatively, \citet{frome2013devise}, \citet{lazaridou2014wampimuk} and \citet{socher2013zero} perform \textit{zero-shot} image classification with an image-to-text neural network mapping. Instead of mapping to latent features, \citet{collell2018acquiring} use a 2-layer feed-forward network to map word embeddings directly to image pixels in order to \textit{visualize} spatial arrangements of objects. Neural networks are also popular in other cross-space applications such as \textit{cross-lingual} tasks. \citet{lazaridou2015hubness} learn a linear map from language A to language B and then translate new words by returning the nearest neighbor of the mapped vector in the B space. 

\begin{figure}[t]
	\begin{center}
		\includegraphics[scale=0.45]{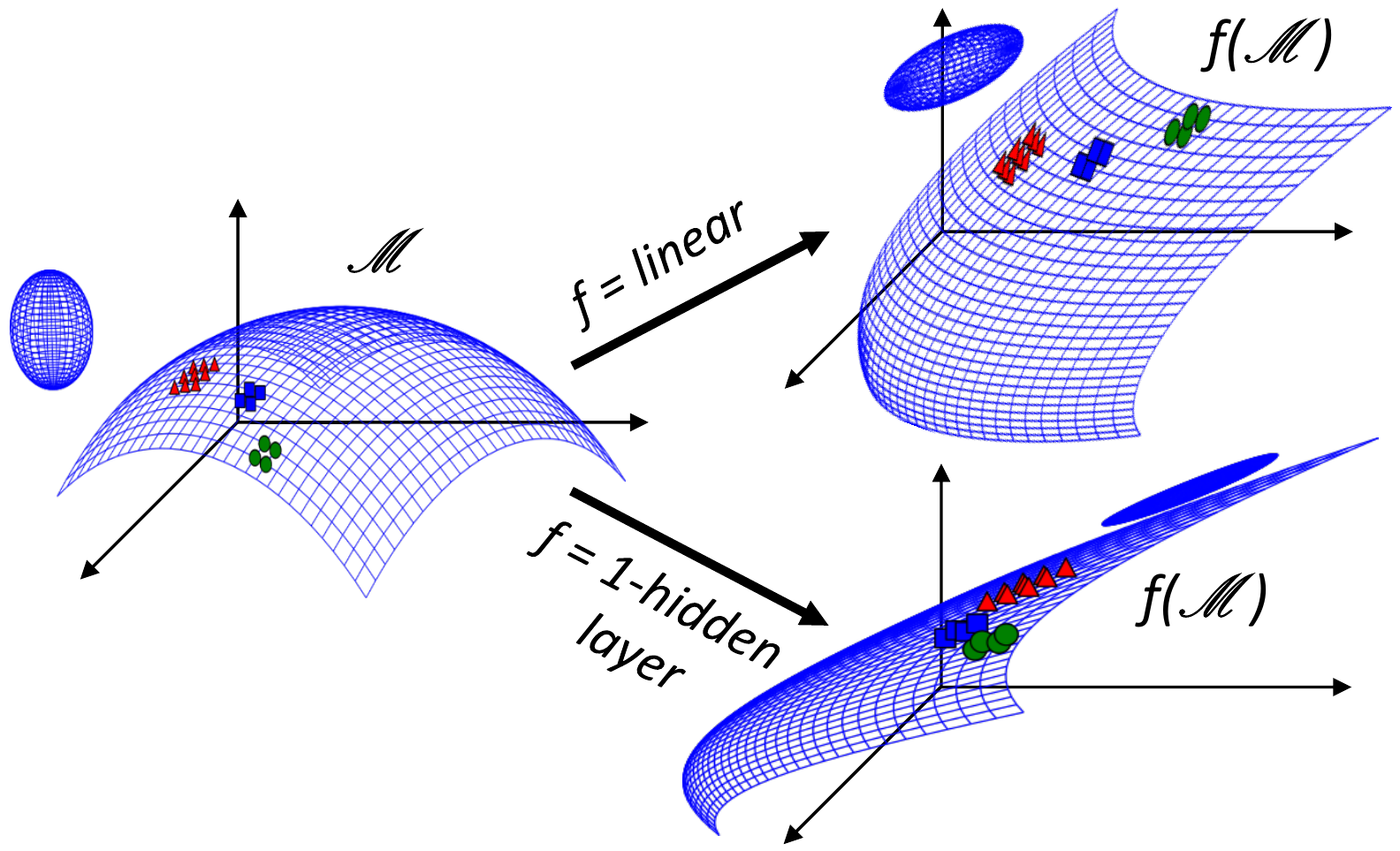}
	\end{center}
	\caption{Effect of applying a mapping $f$ to a (disconnected) manifold $\mathscr{M}$ with three hypothetical classes (\textcolor{blue}{\footnotesize{$\blacksquare$}}, \textcolor{red}{$\blacktriangle$} and \textcolor{green}{\large{$\bullet$}}).}
	\label{fig:rand_mapping}
\end{figure}

In the context of zero-shot learning, shortcomings of cross-space neural mappings have also been identified. For instance, ``hubness" \cite{radovanovic2010existence} and ``pollution" \cite{lazaridou2015hubness} relate to the high-dimensionality of the feature spaces and to overfitting respectively. Crucially, we do not assume that our cross-modal problem has any class labels, and we study the similarity between input and mapped vectors and between output and mapped vectors.

Recent work evidences that the predicted vectors of cross-modal neural net mappings are still largely informative about the input vectors. \citet{lazaridou2015combining} qualitatively observe that abstract textual concepts are grounded with the visual input modality. Counterintuitively, \citet{collell2017imagined} find that the vectors ``imagined" from a language-to-vision neural map, outperform the original visual vectors in concept similarity tasks. The paper argued that the reconstructed visual vectors become grounded with language because the map preserves topological properties of the input. Here, we go one step further and show that the mapped vectors often resemble the input vectors more than the target vectors in semantic terms, which goes against the goal of a cross-modal map.

Well-known theoretical work shows that networks with as few as one hidden layer are able to approximate any function \cite{hornik1989multilayer}. However, this result does not reveal much neither about test performance nor about the semantic structure of the mapped vectors. Instead, the phenomenon described is more closely tied to other properties of neural networks. In particular, continuity guarantees that topological properties of the input, such as connectedness, are preserved \cite{armstrong2013basic}. Furthermore, continuity in a topology induced by a metric also ensures that points that are close together are mapped close together. As a toy example, Fig. \ref{fig:rand_mapping} illustrates the distortion of a manifold after being mapped by a neural net.\footnote{Parameters of these mappings were generated at random.}     

In a noiseless world with fully statistically dependent modalities, the vectors of one modality could be perfectly predicted from those of the other. However, in real-world problems this is unrealistic given the noise of the features and the fact that modalities encode complementary information \cite{collell2016is}. Such unpredictability combined with continuity and topology-preserving properties of neural nets propel the phenomenon identified, namely mapped vectors resembling more the input than the target vectors, in nearest neighbors terms.

\section{Proposed Approach}
\label{approach}

To bridge modalities $\mathcal{X}$ and $\mathcal{Y}$, we consider two popular cross-modal mappings $f:\mathcal{X} \rightarrow \mathcal{Y}$.\\
\\
\noindent{(i) \textbf{Linear} mapping (\textbf{\textit{lin}}):}
\[f(x) =  W_0 x + b_0 \] 
with $W_0 \in \mathbb{R}^{d_y \times d_x}$, $b_0 \in \mathbb{R}^{d_y}$, where $d_x$ and $d_y$ are the input and output dimensions respectively.\\

\noindent{(ii) Feed-forward \textbf{neural network} (\textbf{\textit{nn}}):}
\[ f(x) = W_1 \sigma( W_0 x + b_0) + b_1 \] 
with $W_1 \in \mathbb{R}^{d_y \times d_h}$, $W_0 \in \mathbb{R}^{d_h \times d_x}$, $b_0 \in \mathbb{R}^{d_h}$, $b_1 \in \mathbb{R}^{d_y}$ where $d_h$ is the number of hidden units and $\sigma()$ the non-linearity (e.g., tanh or sigmoid). Although single hidden layer networks are already universal approximators \cite{hornik1989multilayer}, we explored whether deeper nets with \textbf{3 and 5 hidden layers} could improve the fit (see Supplement).\\ 
\\
\noindent{\textbf{Loss}:} Our primary choice is the \textit{MSE}: $\frac{1}{2}\| f(x) - y \|^2$, where $y$ is the target vector. We also tested other losses such as the \textit{cosine}: $1-\cos(f(x), y)$ and the \textit{max-margin}: $\max \{0, \gamma + \|f(x) - y\| - \|f(\tilde{x}) - y\| \}$, where $\tilde{x}$ belongs to a different class than $(x,y)$, and $\gamma$ is the margin. As in \citet{lazaridou2015hubness} and \citet{weston2011wsabie}, we choose the first $\tilde{x}$ that violates the constraint. Notice that losses that do not require class labels such as \textit{MSE} are suitable for a wider, more general set of tasks than discriminative losses (e.g., cross-entropy). In fact, cross-modal retrieval tasks often do not exhibit any class labels. Additionally, our research question concerns the cross-space mapping problem in isolation (independently of class labels).

Let us denote a set of $N$ input and output vectors by $X \in \mathbb{R}^{N \times d_x}$ and $Y \in \mathbb{R}^{N \times d_y}$ respectively. Each input vector $x_i$ is paired to the output vector $y_i$ of the same index ($i=1,\cdots,N$). 
Let us henceforth denote the mapped input vectors by $f(X) \in \mathbb{R}^{N \times d_y}$. In order to explore the similarity between $f(X)$ and $X$, and between $f(X)$ and $Y$, we propose two \textit{ad hoc} settings below.

\subsection{Neighborhood Structure of Mapped Vectors (Experiment 1)}
\label{approach:exp1}

To measure the similarity between the neighborhood structure of two sets of \textit{paired} vectors $V$ and $Z$, we propose the \textit{mean nearest neighbor overlap} measure ($\text{\textbf{\textit{mNNO}}}^K(V,Z)$). We define the \textit{nearest neighbor overlap}  $\text{\textbf{\textit{NNO}}}^K(v_i,z_i)$ as the \textit{number of $K$ nearest neighbors that two paired vectors $v_i, z_i$ share in their respective spaces}. E.g., if the $3$ (= $K$) nearest neighbors of $v_{cat}$ in $V$ are $\{v_{dog}, v_{tiger}, v_{lion}\}$ and those of $z_{cat}$ in $Z$ are $\{z_{mouse}, z_{tiger}, z_{lion}\}$, the $\text{\textbf{\textit{NNO}}}^3(v_{cat},z_{cat})$ is 2. 

\begin{definition}
	Let $V = \{v_i\}^{N}_{i=1}$ and $Z = \{z_i\}^{N}_{i=1}$ be two sets of $N$ paired vectors. We define:
	\begin{equation}
		\text{\textit{\textbf{mNNO}}}^K(V,Z) = \frac{1}{KN} \sum_{i=1}^N \text{NNO}^K(v_i,z_i)
	\end{equation}
	with $\text{\textbf{NNO}}^K(v_i,z_i) = |\text{NN}^K(v_i) \cap \text{NN}^K(z_i)|$, where $\text{NN}^K(v_i)$ and $\text{NN}^K(z_i)$ are the indexes of the $K$ nearest neighbors of $v_i$ and $z_i$, respectively.
\end{definition}

The normalizing constant $K$ simply scales $\text{\textit{mNNO}}^K(V,Z)$ between 0 and 1, making it independent of the choice of $K$. Thus, a $\text{\textit{mNNO}}^K(V,Z)=0.7$ means that the vectors in $V$ and $Z$ share, on average, 70\% of their nearest neighbors.  
Notice that $\text{\textit{mNNO}}$ implicitly performs retrieval for some similarity measure (e.g., Euclidean or cosine), and quantifies how semantically similar two sets of paired vectors are.

\subsection{Mapping with Untrained Networks (Experiment 2)}

To complement the setting above (Sect.~\ref{approach:exp1}), it is instructive to consider the limit case of an untrained network. Concept similarity tasks provide a suitable setting to study the semantic structure of distributed representations \cite{pennington2014glove}. That is, semantically similar concepts should ideally be close together. In particular, our interest is in comparing $X$ with its projection $f(X)$ through a mapping with random parameters, to understand the extent to which the mapping may disrupt or preserve the semantic structure of $X$.

\section{Experimental Setup}
\label{experimental_setup}

\subsection{Experiment 1}

\subsubsection{Datasets}
To test the generality of our claims, we select a rich diversity of cross-modal tasks involving texts at three levels: \textit{word} level (ImageNet), \textit{sentence} level (IAPR TC-12), and \textit{document} level (Wiki).

\noindent{\textbf{ImageNet}} \cite{russakovsky2015imagenet}. Consists of $\sim$14M images, covering $\sim$22K WordNet synsets (or meanings). Following \citet{collell2017imagined}, we take the most relevant word for each synset and keep only synsets with more than 50 images. This yields 9,251 different words (or instances).

\noindent{\textbf{IAPR TC-12}} \cite{grubinger2006iapr}. Contains 20K images (18K train / 2K test) annotated with 255 labels. Each image is accompanied with a short description of one to three sentences.

\noindent{\textbf{Wikipedia}} \cite{pereira2014role}. Has 2,866 samples (2,173 train / 693 test). Each sample is a section of a Wikipedia article paired with one image.

\subsubsection{Hyperparameters and Implementation}
See the Supplement for details.

\subsubsection{Image and Text Features}
\label{text_and_vis_features}
To ensure that results are independent of the choice of image and text features, we use 5 (2 image + 3 text) features of varied dimensionality (64-$d$, 128-$d$, 300-$d$, 2,048-$d$) and two directions, text-to-image ($T\rightarrow I$) and image-to-text ($I\rightarrow T$). We make our extracted features publicly available.\footnote{http://liir.cs.kuleuven.be/software.html}

\noindent{\textbf{Text.}} In \textit{\textbf{ImageNet}} we use 300-dimensional GloVe\footnote{http://nlp.stanford.edu/projects/glove} \cite{pennington2014glove} and 300-$d$ word2vec \cite{Mikolov:2013nips} word embeddings. 
In \textit{\textbf{IAPR TC-12}} and \textit{\textbf{Wiki}}, we employ state-of-the-art bidirectional gated recurrent unit (biGRU) features \cite{cho2014learning} that we learn with a classification task (see Sect. 2 of Supplement).\\
\noindent{\textbf{Image.}} For \textit{\textbf{ImageNet}}, we use the publicly available\footnote{http://liir.cs.kuleuven.be/software.html} VGG-128 \cite{Chatfield14} and ResNet \cite{he2015deep} visual features from \citet{collell2017imagined}, where we obtained 128-dimensional VGG-128 and 2,048-$d$ ResNet features from the last layer (before the softmax) of the forward pass of each image. The final representation for a word is the average feature vector (centroid) of all available images for this word. 
In \textit{\textbf{IAPR TC-12}} and \textit{\textbf{Wiki}}, features for individual images are obtained similarly from the last layer of a ResNet and a VGG-128 model.

\subsection{Experiment 2}

\subsubsection{Datasets}
We include six benchmarks, comprising three types of concept similarity: \textbf{(i) Semantic similarity}: \textit{SemSim} \cite{silberer2014learning}, \textit{Simlex999} \cite{hill2015simlex} and \textit{SimVerb-3500} \cite{gerz2016simverb}; \textbf{(ii) Relatedness}: \textit{MEN} \cite{bruni2014multimodal} and \textit{WordSim-353} \cite{finkelstein2001placing}; \textbf{(iii) Visual similarity}: \textit{VisSim} \cite{silberer2014learning} which includes the same word pairs as \textit{SemSim}, rated for visual similarity instead of semantic. All six test sets contain human ratings of similarity for word pairs, e.g., (`cat',`dog').

\subsubsection{Hyperparameters and Implementation}
The parameters in $W_0, W_1$ are drawn from a random uniform distribution $[-1, 1]$ and $b_0, b_1$ are set to zero. We use a tanh activation $\sigma()$.\footnote{We find that sigmoid and ReLu yield similar results.} The output dimension $d_y$ is set to 2,048 for all embeddings.

\subsubsection{Image and Text Features}
Textual and visual features are the same as described in Sect.~\ref{text_and_vis_features} for the \textit{\textbf{ImageNet}} dataset. 

\subsubsection{Similarity Predictions}
\label{similarity_predictions}
We compute the prediction of similarity between two vectors $z_1, z_2$ with both the cosine $\frac{z_1 z_2}{\| z_1 \|\| z_2 \|}$
and the Euclidean similarity 
$\frac{1}{1+ \| z_1 - z_2 \| }$.\footnote{Notice that papers generally use only cosine similarity \cite{lazaridou2015combining,pennington2014glove}.} 

\subsubsection{Performance Metrics}
As is common practice, we evaluate the predictions of similarity of the embeddings (Sect.~\ref{similarity_predictions}) against the human similarity ratings with the \textit{Spearman correlation} $\rho$. We report the average of 10 sets of randomly generated parameters.

\section{Results and Discussion}
\label{results}

We test statistical significance with a two-sided Wilcoxon rank sum test adjusted with Bonferroni. The null hypothesis is that a compared pair is equal. In Tab.~\ref{tab:exp1}, $^*$ indicates that $\text{\textit{mNNO}}(X,f(X))$ differs from $\text{\textit{mNNO}}(Y,f(X))$ (p $<$ 0.001) on the same mapping, embedding and direction. In Tab.~\ref{tab:exp2}, $^*$ indicates that performance of mapped and input vectors differs (p $<$ 0.05) in the 10 runs.

\subsection{Experiment 1}

Results below are with cosine neighbors and $K$ = 10. Euclidean neighbors yield similar results and are thus left to the Supplement. Similarly, results in ImageNet with GloVe embeddings are shown below and word2vec results in the Supplement. The choice of $K$ = $\{5,10,30\}$ had no visible effect on results. Results with \textit{\textbf{3- and 5-layer}} nets did not show big differences with the results below (see Supplement). The \textit{\textbf{cosine}} and \textit{\textbf{max-margin}} losses performed slightly worse than \textit{MSE} (see Supplement). Although \citet{lazaridou2015hubness} and \citet{weston2011wsabie} find that \textit{max-margin} performs the best in their tasks, we do not find our result entirely surprising given that max-margin focuses on inter-class differences while we look also at intra-class neighbors (in fact, we do not require classes).

Tab.~\ref{tab:exp1} shows our core finding, namely that the semantic structure of $f(X)$ resembles more that of $X$ than that of $Y$, for both \textit{lin} and \textit{nn} maps.

\begin{table}[h!]
	\center
	\resizebox{\columnwidth}{!}{ 
\begin{tabular}{@{}lcccccc@{}}
	\toprule
	&  &  & \multicolumn{2}{c}{ResNet} & \multicolumn{2}{c}{VGG-128} \\ \midrule
	&  &  & $X,f(X)$ & $Y,f(X)$ & $X,f(X)$ & $Y,f(X)$ \\ \midrule
	\multirow{4}{*}{\rotatebox[origin=c]{90}{ImageNet}} & \multirow{2}{*}{$I\rightarrow T$} & \textit{lin} & \textbf{0.681}$^*$ & 0.262 & \textbf{0.723}$^*$ & 0.236 \\
	&  & \textit{nn} & \textbf{0.622}$^*$ & 0.273 & \textbf{0.682}$^*$ & 0.246 \\ \cmidrule(l){2-7} 
	& \multirow{2}{*}{$T\rightarrow I$} & \textit{lin} & \textbf{0.379}$^*$ & 0.241 & \textbf{0.339}$^*$ & 0.229 \\
	&  & \textit{nn} & \textbf{0.354}$^*$ & 0.27 & \textbf{0.326}$^*$ & 0.256 \\ \midrule
	\multirow{4}{*}{\rotatebox[origin=c]{90}{IAPR TC-12}} & \multirow{2}{*}{$I\rightarrow T$} & \textit{lin} & \textbf{0.358}$^*$ & 0.214 & \textbf{0.382}$^*$ & 0.163 \\
	&  & \textit{nn} & \textbf{0.336}$^*$ & 0.219 & \textbf{0.331}$^*$ & 0.18 \\ \cmidrule(l){2-7} 
	& \multirow{2}{*}{$T\rightarrow I$} & \textit{lin} & \textbf{0.48}$^*$ & 0.2 & \textbf{0.419}$^*$ & 0.167 \\
	&  & \textit{nn} & \textbf{0.413}$^*$ & 0.225 & \textbf{0.372}$^*$ & 0.182 \\ \midrule
	\multirow{4}{*}{\rotatebox[origin=c]{90}{Wikipedia}} & \multirow{2}{*}{$I\rightarrow T$} & \textit{lin} & \textbf{0.235}$^*$ & 0.156 & \textbf{0.235}$^*$ & 0.143 \\
	&  & \textit{nn} & \textbf{0.269}$^*$ & 0.161 & \textbf{0.282}$^*$ & 0.148 \\ \cmidrule(l){2-7} 
	& \multirow{2}{*}{$T\rightarrow I$} & \textit{lin} & \textbf{0.574}$^*$ & 0.156 & \textbf{0.6}$^*$ & 0.148 \\
	&  & \textit{nn} & \textbf{0.521}$^*$ & 0.156 & \textbf{0.511}$^*$ & 0.151 \\ \bottomrule
\end{tabular}
	} 
	\caption{Test mean nearest neighbor overlap. Boldface indicates the largest score at each $\text{\textit{mNNO}}^{10}(X,f(X))$ and  $\text{\textit{mNNO}}^{10}(Y,f(X))$ pair, which are abbreviated by $X,f(X)$ and $Y,f(X)$.}
	\label{tab:exp1}
\end{table}

\begin{figure}[t]
	\begin{center}
		\includegraphics[scale=0.3]{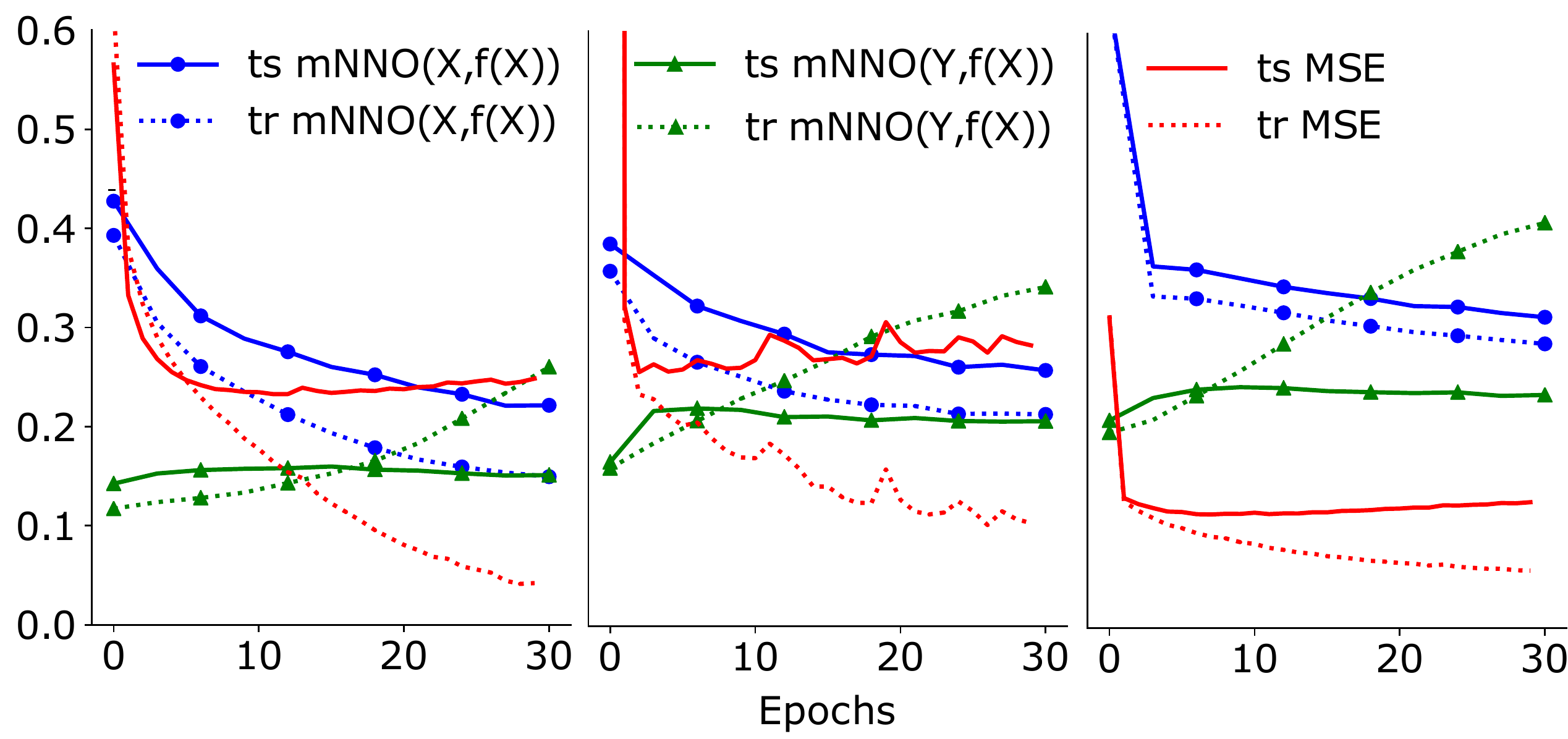}
	\end{center}
	\caption{Learning a \textit{\textbf{nn}} model 
		in \textbf{Wiki} (left), \textbf{IAPR TC-12} (middle) and \textbf{ImageNet} (right).}
	\label{fig:epochs}
\end{figure}

Fig. \ref{fig:epochs} is particularly revealing. If we would only look at \textit{train} performance (and allow train MSE to reach 0) then $f(X)=Y$ and clearly train $\text{\textit{mNNO}}(f(X),Y)=1$ while $\text{\textit{mNNO}}(f(X),X)$ can only be smaller than 1. However, the interest is always on \textit{test} samples, and (near-)perfect \textit{test} prediction is unrealistic. Notice in fact in Fig.~\ref{fig:epochs} that even if we look at \textit{train} fit, MSE needs to be close to 0 for $\text{\textit{mNNO}}(f(X),Y)$ to be reasonably large. In all the combinations from Tab.~\ref{tab:exp1}, the \textit{test} $\text{\textit{mNNO}}(f(X),Y)$ never surpasses \textit{test} $\text{\textit{mNNO}}(f(X),X)$ for any number of epochs, even with an oracle (not shown).

\subsection{Experiment 2}

Tab.~\ref{tab:exp2} shows that untrained linear ($f_{\text{lin}}$) and neural net ($f_{\text{nn}}$) mappings preserve the semantic structure of the input $X$, complementing thus the findings of Experiment 1. Experiment 1 concerns learning, while, by ``ablating" the learning part and randomizing weights, Experiment 2 is revealing about the natural tendency of neural nets to preserve semantic information about the input, regardless of the choice of the target vectors and  loss function.

\begin{table}[h!]
	\center
	\resizebox{\columnwidth}{!}{ 
\begin{tabular}{@{}lcccccc@{}}
	\toprule
	& \multicolumn{2}{c}{WS-353} & \multicolumn{2}{c}{Men} & \multicolumn{2}{c}{SemSim} \\ \cmidrule(l){1-7} 
	& Cos & Eucl & Cos & Eucl & Cos & Eucl \\ \midrule
$f_{\text{nn}}$(GloVe) & \textbf{0.632} & \textbf{0.634}$^*$ & 0.795 & \textbf{0.791}$^*$ & 0.75$^*$ & \textbf{0.744}$^*$ \\
$f_{\text{lin}}$(GloVe) & 0.63 & 0.606 & 0.798 & 0.781 & 0.763 & 0.712 \\
GloVe & \textbf{0.632} & 0.601 & \textbf{0.801} & 0.782 & \textbf{0.768} & 0.716 \\ \midrule
$f_{\text{nn}}$(ResNet) & 0.402 & 0.408$^*$ & 0.556 & \textbf{0.554}$^*$ & 0.512 & 0.513 \\
$f_{\text{lin}}$(ResNet) & \textbf{0.425} & 0.449 & 0.566 & 0.534 & 0.533 & 0.514 \\
ResNet & 0.423 & \textbf{0.457} & \textbf{0.567} & 0.535 & \textbf{0.534} & \textbf{0.516} \\ \midrule
& \multicolumn{2}{c}{VisSim} & \multicolumn{2}{c}{SimLex} & \multicolumn{2}{c}{SimVerb} \\ \cmidrule(l){1-7} 
& Cos & Eucl & Cos & Eucl & Cos & Eucl \\ \midrule
$f_{\text{nn}}$(GloVe) & 0.594$^*$ & \textbf{0.59}$^*$ & 0.369 & \textbf{0.363}$^*$ & 0.313 & \textbf{0.301}$^*$ \\
$f_{\text{lin}}$(GloVe) & 0.602$^*$ & 0.576 & 0.369 & 0.341 & \textbf{0.326} & 0.23 \\
GloVe & \textbf{0.606} & 0.58 & \textbf{0.371} & 0.34 & 0.32 & 0.235 \\ \midrule
$f_{\text{nn}}$(ResNet) & 0.527$^*$ & \textbf{0.526}$^*$ & 0.405 & \textbf{0.406} & 0.178 & 0.169 \\
$f_{\text{lin}}$(ResNet) & 0.541 & 0.498 & \textbf{0.409} & 0.404 & 0.198 & 0.182 \\
ResNet & \textbf{0.543} & 0.501 & \textbf{0.409} & 0.403 & \textbf{0.211} & \textbf{0.199} \\ \bottomrule
\end{tabular}
	} 
		\caption{Spearman correlations between human ratings and the similarities (cosine or Euclidean) predicted from the embeddings. Boldface denotes best performance per input embedding type.}
		\label{tab:exp2}
\end{table}

\section{Conclusions}
\label{conclusions}

Overall, we uncovered a phenomenon neglected so far, namely that neural net cross-modal mappings can produce mapped vectors more akin to the input vectors than the target vectors, in terms of semantic structure. Such finding has been possible thanks to the proposed measure that explicitly quantifies similarity between the neighborhood structure of two sets of vectors. While other measures such as mean squared error can be misleading, our measure provides a more realistic estimate of the semantic similarity between predicted and target vectors. In fact, it is the semantic structure (or pairwise similarities) what ultimately matters in cross-modal applications.

\section*{Acknowledgments}
This work has been supported by the CHIST-ERA EU project MUSTER\footnote{http://www.chistera.eu/projects/muster} and by the KU Leuven grant RUN/15/005.

\bibliography{cross_modal_mapping}

\begin{thebibliography}{}
\expandafter\ifx\csname natexlab\endcsname\relax\def\natexlab#1{#1}\fi

\bibitem[{Armstrong(2013)}]{armstrong2013basic}
Mark~Anthony Armstrong. 2013.
\newblock {\em Basic topology\/}.
\newblock Springer Science \& Business Media.

\bibitem[{Bruni et~al.(2014)Bruni, Tran, and Baroni}]{bruni2014multimodal}
Elia Bruni, Nam-Khanh Tran, and Marco Baroni. 2014.
\newblock Multimodal distributional semantics.
\newblock {\em JAIR\/} 49(1-47).

\bibitem[{Chatfield et~al.(2014)Chatfield, Simonyan, Vedaldi, and
  Zisserman}]{Chatfield14}
Ken Chatfield, Karen Simonyan, Andrea Vedaldi, and Andrew Zisserman. 2014.
\newblock Return of the devil in the details: Delving deep into convolutional
  nets.
\newblock In {\em BMVC\/}.

\bibitem[{Cho et~al.(2014)Cho, Van~Merri{\"e}nboer, Gulcehre, Bahdanau,
  Bougares, Schwenk, and Bengio}]{cho2014learning}
Kyunghyun Cho, Bart Van~Merri{\"e}nboer, Caglar Gulcehre, Dzmitry Bahdanau,
  Fethi Bougares, Holger Schwenk, and Yoshua Bengio. 2014.
\newblock Learning phrase representations using rnn encoder-decoder for
  statistical machine translation.
\newblock {\em arXiv preprint arXiv:1406.1078\/} .

\bibitem[{Chollet et~al.(2015)}]{chollet2015keras}
Fran\c{c}ois Chollet et~al. 2015.
\newblock Keras.
\newblock \url{https://github.com/keras-team/keras}.

\bibitem[{Collell and Moens(2016)}]{collell2016is}
Guillem Collell and Marie-Francine Moens. 2016.
\newblock {Is an Image Worth More than a Thousand Words? On the Fine-Grain
  Semantic Differences between Visual and Linguistic Representations}.
\newblock In {\em COLING\/}. ACL, pages 2807--2817.

\bibitem[{Collell et~al.(2018)Collell, Van~Gool, and
  Moens}]{collell2018acquiring}
Guillem Collell, Luc Van~Gool, and Marie-Francine Moens. 2018.
\newblock {Acquiring Common Sense Spatial Knowledge through Implicit Spatial
  Templates}.
\newblock In {\em AAAI\/}. AAAI.

\bibitem[{Collell et~al.(2017)Collell, Zhang, and Moens}]{collell2017imagined}
Guillem Collell, Teddy Zhang, and Marie-Francine Moens. 2017.
\newblock {Imagined Visual Representations as Multimodal Embeddings}.
\newblock In {\em AAAI\/}. AAAI, pages 4378--4384.

\bibitem[{Finkelstein et~al.(2001)Finkelstein, Gabrilovich, Matias, Rivlin,
  Solan, Wolfman, and Ruppin}]{finkelstein2001placing}
Lev Finkelstein, Evgeniy Gabrilovich, Yossi Matias, Ehud Rivlin, Zach Solan,
  Gadi Wolfman, and Eytan Ruppin. 2001.
\newblock Placing search in context: The concept revisited.
\newblock In {\em WWW\/}. ACM, pages 406--414.

\bibitem[{Frome et~al.(2013)Frome, Corrado, Shlens, Bengio, Dean, Mikolov
  et~al.}]{frome2013devise}
Andrea Frome, Greg~S Corrado, Jon Shlens, Samy Bengio, Jeff Dean, Tomas
  Mikolov, et~al. 2013.
\newblock Devise: A deep visual-semantic embedding model.
\newblock In {\em NIPS\/}. pages 2121--2129.

\bibitem[{Gerz et~al.(2016)Gerz, Vuli{\'c}, Hill, Reichart, and
  Korhonen}]{gerz2016simverb}
Daniela Gerz, Ivan Vuli{\'c}, Felix Hill, Roi Reichart, and Anna Korhonen.
  2016.
\newblock Simverb-3500: A large-scale evaluation set of verb similarity.
\newblock {\em arXiv preprint arXiv:1608.00869\/} .

\bibitem[{Grubinger et~al.(2006)Grubinger, Clough, M{\"u}ller, and
  Deselaers}]{grubinger2006iapr}
Michael Grubinger, Paul Clough, Henning M{\"u}ller, and Thomas Deselaers. 2006.
\newblock The iapr tc-12 benchmark: A new evaluation resource for visual
  information systems.
\newblock In {\em International workshop ontoImage\/}. volume~5, page~10.

\bibitem[{He et~al.(2015)He, Zhang, Ren, and Sun}]{he2015deep}
Kaiming He, Xiangyu Zhang, Shaoqing Ren, and Jian Sun. 2015.
\newblock Deep residual learning for image recognition.
\newblock {\em arXiv preprint arXiv:1512.03385\/} .

\bibitem[{Hill et~al.(2015)Hill, Reichart, and Korhonen}]{hill2015simlex}
Felix Hill, Roi Reichart, and Anna Korhonen. 2015.
\newblock Simlex-999: Evaluating semantic models with (genuine) similarity
  estimation.
\newblock {\em Computational Linguistics\/} 41(4):665--695.

\bibitem[{Hornik et~al.(1989)Hornik, Stinchcombe, and
  White}]{hornik1989multilayer}
Kurt Hornik, Maxwell Stinchcombe, and Halbert White. 1989.
\newblock Multilayer feedforward networks are universal approximators.
\newblock {\em Neural networks\/} 2(5):359--366.

\bibitem[{Lazaridou et~al.(2014)Lazaridou, Bruni, and
  Baroni}]{lazaridou2014wampimuk}
Angeliki Lazaridou, Elia Bruni, and Marco Baroni. 2014.
\newblock Is this a wampimuk? cross-modal mapping between distributional
  semantics and the visual world.
\newblock In {\em ACL\/}. pages 1403--1414.

\bibitem[{Lazaridou et~al.(2015{\natexlab{a}})Lazaridou, Dinu, and
  Baroni}]{lazaridou2015hubness}
Angeliki Lazaridou, Georgiana Dinu, and Marco Baroni. 2015{\natexlab{a}}.
\newblock Hubness and pollution: Delving into cross-space mapping for zero-shot
  learning.
\newblock In {\em ACL\/}. volume~1, pages 270--280.

\bibitem[{Lazaridou et~al.(2015{\natexlab{b}})Lazaridou, Pham, and
  Baroni}]{lazaridou2015combining}
Angeliki Lazaridou, Nghia~The Pham, and Marco Baroni. 2015{\natexlab{b}}.
\newblock Combining language and vision with a multimodal skip-gram model.
\newblock {\em arXiv preprint arXiv:1501.02598\/} .

\bibitem[{Mikolov et~al.(2013)Mikolov, Sutskever, Chen, Corrado, and
  Dean}]{Mikolov:2013nips}
Tomas Mikolov, Ilya Sutskever, Kai Chen, Gregory~S. Corrado, and Jeffrey Dean.
  2013.
\newblock Distributed representations of words and phrases and their
  compositionality.
\newblock In {\em NIPS\/}. pages 3111--3119.

\bibitem[{Pennington et~al.(2014)Pennington, Socher, and
  Manning}]{pennington2014glove}
Jeffrey Pennington, Richard Socher, and Christopher~D Manning. 2014.
\newblock Glove: Global vectors for word representation.
\newblock In {\em EMNLP\/}. volume~14, pages 1532--1543.

\bibitem[{Pereira et~al.(2014)Pereira, Coviello, Doyle, Rasiwasia, Lanckriet,
  Levy, and Vasconcelos}]{pereira2014role}
Jose~Costa Pereira, Emanuele Coviello, Gabriel Doyle, Nikhil Rasiwasia, Gert~RG
  Lanckriet, Roger Levy, and Nuno Vasconcelos. 2014.
\newblock On the role of correlation and abstraction in cross-modal multimedia
  retrieval.
\newblock {\em TPAMI\/} 36(3):521--535.

\bibitem[{Qiao et~al.(2017)Qiao, Liu, Shen, and Hengel}]{qiao2017visually}
Ruizhi Qiao, Lingqiao Liu, Chunhua Shen, and Anton van~den Hengel. 2017.
\newblock Visually aligned word embeddings for improving zero-shot learning.
\newblock {\em arXiv preprint arXiv:1707.05427\/} .

\bibitem[{Radovanovi{\'c} et~al.(2010)Radovanovi{\'c}, Nanopoulos, and
  Ivanovi{\'c}}]{radovanovic2010existence}
Milos Radovanovi{\'c}, Alexandros Nanopoulos, and Mirjana Ivanovi{\'c}. 2010.
\newblock On the existence of obstinate results in vector space models.
\newblock In {\em SIGIR\/}. ACM, pages 186--193.

\bibitem[{Russakovsky et~al.(2015)Russakovsky, Deng, Su, Krause, Satheesh, Ma,
  Huang, Karpathy, Khosla, Bernstein et~al.}]{russakovsky2015imagenet}
Olga Russakovsky, Jia Deng, Hao Su, Jonathan Krause, Sanjeev Satheesh, Sean Ma,
  Zhiheng Huang, Andrej Karpathy, Aditya Khosla, Michael Bernstein, et~al.
  2015.
\newblock Imagenet large scale visual recognition challenge.
\newblock {\em IJCV\/} 115(3):211--252.

\bibitem[{Silberer and Lapata(2014)}]{silberer2014learning}
Carina Silberer and Mirella Lapata. 2014.
\newblock Learning grounded meaning representations with autoencoders.
\newblock In {\em ACL\/}. pages 721--732.

\bibitem[{Socher et~al.(2013)Socher, Ganjoo, Manning, and Ng}]{socher2013zero}
Richard Socher, Milind Ganjoo, Christopher~D Manning, and Andrew Ng. 2013.
\newblock Zero-shot learning through cross-modal transfer.
\newblock In {\em NIPS\/}. pages 935--943.

\bibitem[{Wang et~al.(2016)Wang, Yin, Wang, Wu, and
  Wang}]{wang2016comprehensive}
Kaiye Wang, Qiyue Yin, Wei Wang, Shu Wu, and Liang Wang. 2016.
\newblock A comprehensive survey on cross-modal retrieval.
\newblock {\em arXiv preprint arXiv:1607.06215\/} .

\bibitem[{Weston et~al.(2011)Weston, Bengio, and Usunier}]{weston2011wsabie}
Jason Weston, Samy Bengio, and Nicolas Usunier. 2011.
\newblock Wsabie: Scaling up to large vocabulary image annotation.
\newblock In {\em IJCAI\/}. volume~11, pages 2764--2770.

\bibitem[{Zhang et~al.(2016)Zhang, Gong, and Shah}]{zhang2016fast}
Yang Zhang, Boqing Gong, and Mubarak Shah. 2016.
\newblock Fast zero-shot image tagging.
\newblock In {\em CVPR\/}. IEEE, pages 5985--5994.

\end{thebibliography}
\bibliographystyle{acl_natbib}

\appendix

\onecolumn

\begin{center}
	\begin{huge} Supplementary Material of: \end{huge}\\
	\LARGE{Do Neural Network Cross-Modal Mappings Really Bridge Modalities?}
\end{center}

\section{Hyperparameters and Implementation}

Hyperparameters (including number of epochs) are chosen by 5 fold cross-validation (CV) optimizing for the test loss. Crucially, we ensure that all mappings are learned properly by verifying that the training loss steadily decreases. We search learning rates in \{0.01, 0.001, 0.0001\} and number of hidden units ($d_h$) in \{64, 128, 256, 512, 1024\}. 

Using different number of hidden units (and selecting the best-performing one) is important in order to guarantee that our conclusions are not influenced or just a product of underfitting or overfitting. Similarly, we learned the mappings at different levels of dropout \{0.25, 0.5, 0.75\} which did not yield any improvement w.r.t. zero dropout (shown in our results). 

We use a ReLu activation, the RMSprop optimizer ($\rho = 0.9$, $\epsilon = 10^{-8}$) and a batch size of 64. We find that sigmoid and tanh yield similar results as ReLu. Our implementation is in Keras \citep{chollet2015keras}.

Since ImageNet does not have any set of ``test concepts", we employ 5-fold CV. Reported results are either averages on 5 folds (ImageNet) or 5 runs with different model weights initializations (IAPR TC-12 and Wiki).

For the \textit{max-margin} loss, we choose the margin $\gamma$ by cross-validation and explore values within $\{1, 2.5, 5, 7.5, 10\}$.

\section{Textual Feature Extraction}
Unlike ImageNet where we associate a word embedding to each concept, the textual modality in IAPR TC-12 and Wiki consists of sentences. In order to extract state-of-the art textual features in these datasets we train the following, separate network (prior to the cross-modal mapping). First, the embedded input sentences are passed to a bidirectional GRU of 64 units, then fed into a fully-connected layer, followed by a cross-entropy loss on the vector of class labels. We collect the 64-d averaged GRU hidden states of both directions as features. The network is trained with the Adam optimizer. 

In Wiki and IAPR TC-12 we verify that the extracted text and image features are indeed informative and useful by computing their mean average precision (mAP) in retrieval (considering that a document B is relevant for document A if A and B share at least one class label). In Wiki we find mAPs of: biGRU = 0.77, ResNet = 0.22 and vgg128 = 0.21. In IAPR TC-12 we find mAPs of: biGRU = 0.77, ResNet = 0.49 and vgg128 = 0.46. Notice that ImageNet has a single data point per class in our setting, and thus mAP cannot be computed. However, we employ standard GloVe, word2vec, VGG-128 and ResNet vectors in ImageNet, which are known to perform well.

\section{Additional Results}
\label{sec:supplemental}

\paragraph{Results with $\text{\textit{mNNO}}(X,Y)$} (omitted in the main paper for space reasons): Interestingly, the similarity $\text{\textit{mNNO}}(X,Y)$ between original input $X$ and output $Y$ vectors is generally low (between 1.5 and 2.3), indicating that these spaces are originally quite different. However, $\text{\textit{mNNO}}(X,Y)$ always remains lower than $\text{\textit{mNNO}}(f(X),Y)$, indicating thus that the mapping makes a difference.

\subsection{Experiment 1}

\subsubsection{Results with 3 and 5 layers}

\begin{table}[H]
	\center
	\begin{tabular}{lcccccc}
		\hline
		&  &  & \multicolumn{2}{c}{ResNet} & \multicolumn{2}{c}{VGG-128} \\ \hline
		&  &  & $X,f(X)$ & $Y,f(X)$ & $X,f(X)$ & $Y,f(X)$ \\ \hline
		\multirow{4}{*}{\rotatebox[origin=c]{90}{ImageNet}} & \multirow{2}{*}{$I\rightarrow T$} & nn-3 & \textbf{0.571} & 0.279 & \textbf{0.602} & 0.258 \\
		&  & nn-5 & \textbf{0.615} & 0.275 & \textbf{0.644} & 0.255 \\ \cline{2-7} 
		& \multirow{2}{*}{$T\rightarrow I$} & nn-3 & \textbf{0.274} & 0.27 & 0.254 & \textbf{0.256} \\
		&  & nn-5 & \textbf{0.286} & 0.274 & \textbf{0.273} & 0.259 \\ \hline
		\multirow{4}{*}{\rotatebox[origin=c]{90}{IAPR TC}} & \multirow{2}{*}{$I\rightarrow T$} & nn-3 & \textbf{0.301} & 0.225 & \textbf{0.288} & 0.181 \\
		&  & nn-5 & \textbf{0.29} & 0.227 & \textbf{0.308} & 0.184 \\ \cline{2-7} 
		& \multirow{2}{*}{$T\rightarrow I$} & nn-3 & \textbf{0.324} & 0.229 & \textbf{0.294} & 0.18 \\
		&  & nn-5 & \textbf{0.355} & 0.232 & \textbf{0.339} & 0.183 \\ \hline
		\multirow{4}{*}{\rotatebox[origin=c]{90}{Wiki}} & \multirow{2}{*}{$I\rightarrow T$} & nn-3 & \textbf{0.227} & 0.159 & \textbf{0.247} & 0.144 \\
		&  & nn-5 & \textbf{0.275} & 0.163 & \textbf{0.262} & 0.146 \\ \cline{2-7} 
		& \multirow{2}{*}{$T\rightarrow I$} & nn-3 & \textbf{0.367} & 0.148 & \textbf{0.342} & 0.145 \\
		&  & nn-5 & \textbf{0.412} & 0.152 & \textbf{0.428} & 0.147 \\ \hline
	\end{tabular}
	\caption{Test mean nearest neighbor overlap with 3- and 5-hidden layer neural networks, using cosine-based neighbors and MSE loss. Boldface indicates best performance between each $\text{\textit{mNNO}}^{10}(X,f(X))$ and  $\text{\textit{mNNO}}^{10}(Y,f(X))$ pair, which are abbreviated by $X,f(X)$ and $Y,f(X)$.}
	\label{tab:exp1_3_5_layers_cosine}
\end{table}

It is interesting to notice that even though the difference between $\text{\textit{mNNO}}^{10}(X,f(X))$ and  $\text{\textit{mNNO}}^{10}(Y,f(X))$ has narrowed down w.r.t. the linear and 1-hidden layer models (in the main paper) in some cases (e.g., ImageNet), this does not seem to be caused by better predictions, i.e., an increase of $\text{\textit{mNNO}}^{10}(Y,f(X))$, but rather by a decrease of $\text{\textit{mNNO}}^{10}(X,f(X))$. This is expected since with more layers the information about the input is less preserved.

\begin{table}[H]
	\center
	\begin{tabular}{lcccccc}
		\hline
		&  &  & \multicolumn{2}{c}{ResNet} & \multicolumn{2}{c}{VGG-128} \\ \hline
		&  &  & $X,f(X)$ & $Y,f(X)$ & $X,f(X)$ & $Y,f(X)$ \\ \hline
		\multirow{4}{*}{\rotatebox[origin=c]{90}{ImageNet}} & \multirow{2}{*}{$I\rightarrow T$} & nn-3 & \textbf{0.562} & 0.243 & \textbf{0.574} & 0.229 \\
		&  & nn-5 & \textbf{0.61} & 0.241 & \textbf{0.619} & 0.227 \\ \cline{2-7} 
		& \multirow{2}{*}{$T\rightarrow I$} & nn-3 & 0.252 & \textbf{0.263} & 0.23 & \textbf{0.244} \\
		&  & nn-5 & 0.261 & \textbf{0.264} & \textbf{0.243} & 0.242 \\ \hline
		\multirow{4}{*}{\rotatebox[origin=c]{90}{IAPR TC}} & \multirow{2}{*}{$I\rightarrow T$} & nn-3 & \textbf{0.275} & 0.208 & \textbf{0.259} & 0.174 \\
		&  & nn-5 & \textbf{0.262} & 0.207 & \textbf{0.276} & 0.174 \\ \cline{2-7} 
		& \multirow{2}{*}{$T\rightarrow I$} & nn-3 & \textbf{0.312} & 0.215 & \textbf{0.27} & 0.168 \\
		&  & nn-5 & \textbf{0.351} & 0.218 & \textbf{0.315} & 0.17 \\ \hline
		\multirow{4}{*}{\rotatebox[origin=c]{90}{Wiki}} & \multirow{2}{*}{$I\rightarrow T$} & nn-3 & \textbf{0.219} & 0.15 & \textbf{0.239} & 0.14 \\
		&  & nn-5 & \textbf{0.259} & 0.152 & \textbf{0.25} & 0.143 \\ \cline{2-7} 
		& \multirow{2}{*}{$T\rightarrow I$} & nn-3 & \textbf{0.375} & 0.145 & \textbf{0.363} & 0.134 \\
		&  & nn-5 & \textbf{0.431} & 0.144 & \textbf{0.426} & 0.135 \\ \hline
	\end{tabular}
	\caption{Test mean nearest neighbor overlap with 3- and 5-hidden layer neural networks, using Euclidean neighbors and MSE loss.}
	\label{tab:exp1_3_5_layers_euclidean}
\end{table}

\subsubsection{Results with the max margin loss}

\begin{table}[H]
	\center
	\begin{tabular}{lcccccc}
		\hline
		&  &  & \multicolumn{2}{c}{ResNet} & \multicolumn{2}{c}{VGG-128} \\ \hline
		&  &  & $X,f(X)$ & $Y,f(X)$ & $X,f(X)$ & $Y,f(X)$ \\ \hline
		\multirow{4}{*}{\rotatebox[origin=c]{90}{ImageNet}} & \multirow{2}{*}{$I\rightarrow T$} & lin & \textbf{0.739} & 0.253 & \textbf{0.779} & 0.235 \\
		&  & nn & \textbf{0.769} & 0.233 & \textbf{0.736} & 0.238 \\ \cline{2-7} 
		& \multirow{2}{*}{$T\rightarrow I$} & lin & \textbf{0.526} & 0.252 & \textbf{0.454} & 0.241 \\
		&  & nn & \textbf{0.419} & 0.23 & \textbf{0.378} & 0.22 \\ \hline
		\multirow{4}{*}{\rotatebox[origin=c]{90}{IAPR TC}} & \multirow{2}{*}{$I\rightarrow T$} & lin & \textbf{0.423} & 0.205 & \textbf{0.441} & 0.164 \\
		&  & nn & \textbf{0.291} & 0.179 & \textbf{0.36} & 0.16 \\ \cline{2-7} 
		& \multirow{2}{*}{$T\rightarrow I$} & lin & \textbf{0.674} & 0.198 & \textbf{0.604} & 0.17 \\
		&  & nn & \textbf{0.592} & 0.215 & \textbf{0.529} & 0.176 \\ \hline
		\multirow{4}{*}{\rotatebox[origin=c]{90}{Wiki}} & \multirow{2}{*}{$I\rightarrow T$} & lin & \textbf{0.366} & 0.156 & \textbf{0.333} & 0.152 \\
		&  & nn & \textbf{0.236} & 0.153 & \textbf{0.399} & 0.153 \\ \cline{2-7} 
		& \multirow{2}{*}{$T\rightarrow I$} & lin & \textbf{0.725} & 0.151 & \textbf{0.723} & 0.146 \\
		&  & nn & \textbf{0.701} & 0.151 & \textbf{0.662} & 0.146 \\ \hline
	\end{tabular}
	\caption{Test mean nearest neighbor overlap with cosine-based neighbors and \textit{max margin loss}.}
	\label{tab:exp1_max-margin_loss_cosine}
\end{table}

\begin{table}[H]
	\center
	\begin{tabular}{lcccccc}
		\hline
		&  &  & \multicolumn{2}{c}{ResNet} & \multicolumn{2}{c}{VGG-128} \\ \hline
		&  &  & $X,f(X)$ & $Y,f(X)$ & $X,f(X)$ & $Y,f(X)$ \\ \hline
		\multirow{4}{*}{\rotatebox[origin=c]{90}{ImageNet}} & \multirow{2}{*}{$I\rightarrow T$} & lin & \textbf{0.762} & 0.229 & \textbf{0.776} & 0.209 \\
		&  & nn & \textbf{0.776} & 0.213 & \textbf{0.724} & 0.214 \\ \cline{2-7} 
		& \multirow{2}{*}{$T\rightarrow I$} & lin & \textbf{0.49} & 0.241 & \textbf{0.418} & 0.225 \\
		&  & nn & \textbf{0.384} & 0.221 & \textbf{0.343} & 0.212 \\ \hline
		\multirow{4}{*}{\rotatebox[origin=c]{90}{IAPR TC}} & \multirow{2}{*}{$I\rightarrow T$} & lin & \textbf{0.409} & 0.195 & \textbf{0.447} & 0.155 \\
		&  & nn & \textbf{0.275} & 0.172 & \textbf{0.329} & 0.15 \\ \cline{2-7} 
		& \multirow{2}{*}{$T\rightarrow I$} & lin & \textbf{0.685} & 0.189 & \textbf{0.619} & 0.158 \\
		&  & nn & \textbf{0.558} & 0.201 & \textbf{0.49} & 0.162 \\ \hline
		\multirow{4}{*}{\rotatebox[origin=c]{90}{Wiki}} & \multirow{2}{*}{$I\rightarrow T$} & lin & \textbf{0.38} & 0.154 & \textbf{0.339} & 0.142 \\
		&  & nn & \textbf{0.232} & 0.144 & \textbf{0.398} & 0.141 \\ \cline{2-7} 
		& \multirow{2}{*}{$T\rightarrow I$} & lin & \textbf{0.789} & 0.143 & \textbf{0.773} & 0.135 \\
		&  & nn & \textbf{0.724} & 0.14 & \textbf{0.723} & 0.135 \\ \hline
	\end{tabular}
	\caption{Test mean nearest neighbor overlap with Euclidean-based neighbors and \textit{max margin loss}.}
	\label{tab:exp1_max-margin_loss_euclidean}
\end{table}

\subsubsection{Results with the cosine loss}

\begin{table}[H]
	\center
	\begin{tabular}{lcccccc}
		\hline
		&  &  & \multicolumn{2}{c}{ResNet} & \multicolumn{2}{c}{VGG-128} \\ \hline
		&  &  & $X,f(X)$ & $Y,f(X)$ & $X,f(X)$ & $Y,f(X)$ \\ \hline
		\multirow{4}{*}{\rotatebox[origin=c]{90}{ImageNet}} & \multirow{2}{*}{$I\rightarrow T$} & lin & \textbf{0.697} & 0.268 & \textbf{0.812} & 0.244 \\
		&  & nn & \textbf{0.58} & 0.28 & \textbf{0.629} & 0.256 \\ \cline{2-7} 
		& \multirow{2}{*}{$T\rightarrow I$} & lin & \textbf{0.382} & 0.241 & \textbf{0.336} & 0.224 \\
		&  & nn & \textbf{0.346} & 0.277 & \textbf{0.331} & 0.237 \\ \hline
		\multirow{4}{*}{\rotatebox[origin=c]{90}{IAPR TC}} & \multirow{2}{*}{$I\rightarrow T$} & lin & \textbf{0.37} & 0.213 & \textbf{0.594} & 0.162 \\
		&  & nn & \textbf{0.35} & 0.234 & \textbf{0.516} & 0.158 \\ \cline{2-7} 
		& \multirow{2}{*}{$T\rightarrow I$} & lin & \textbf{0.469} & 0.205 & \textbf{0.405} & 0.169 \\
		&  & nn & \textbf{0.386} & 0.226 & \textbf{0.338} & 0.185 \\ \hline
		\multirow{4}{*}{\rotatebox[origin=c]{90}{Wiki}} & \multirow{2}{*}{$I\rightarrow T$} & lin & \textbf{0.26} & 0.157 & \textbf{0.621} & 0.143 \\
		&  & nn & \textbf{0.213} & 0.156 & \textbf{0.281} & 0.15 \\ \cline{2-7} 
		& \multirow{2}{*}{$T\rightarrow I$} & lin & \textbf{0.549} & 0.157 & \textbf{0.53} & 0.154 \\
		&  & nn & \textbf{0.642} & 0.151 & \textbf{0.547} & 0.149 \\ \hline
	\end{tabular}
	\caption{Test mean nearest neighbor overlap with cosine-based neighbors and \textit{cosine loss}.}
	\label{tab:exp1_cosine_loss_cosine}
\end{table}

\begin{table}[H]
	\center
	\begin{tabular}{lcccccc}
		\hline
		&  &  & \multicolumn{2}{c}{ResNet} & \multicolumn{2}{c}{VGG-128} \\ \hline
		&  &  & $X,f(X)$ & $Y,f(X)$ & $X,f(X)$ & $Y,f(X)$ \\ \hline
		\multirow{4}{*}{\rotatebox[origin=c]{90}{ImageNet}} & \multirow{2}{*}{$I\rightarrow T$} & lin & \textbf{0.698} & 0.236 & \textbf{0.812} & 0.218 \\
		&  & nn & \textbf{0.562} & 0.238 & \textbf{0.597} & 0.218 \\ \cline{2-7} 
		& \multirow{2}{*}{$T\rightarrow I$} & lin & \textbf{0.36} & 0.225 & \textbf{0.319} & 0.209 \\
		&  & nn & \textbf{0.28} & 0.221 & \textbf{0.288} & 0.205 \\ \hline
		\multirow{4}{*}{\rotatebox[origin=c]{90}{IAPR TC}} & \multirow{2}{*}{$I\rightarrow T$} & lin & \textbf{0.351} & 0.197 & \textbf{0.596} & 0.152 \\
		&  & nn & \textbf{0.295} & 0.201 & \textbf{0.452} & 0.144 \\ \cline{2-7} 
		& \multirow{2}{*}{$T\rightarrow I$} & lin & \textbf{0.475} & 0.184 & \textbf{0.409} & 0.153 \\
		&  & nn & \textbf{0.359} & 0.193 & \textbf{0.29} & 0.158 \\ \hline
		\multirow{4}{*}{\rotatebox[origin=c]{90}{Wiki}} & \multirow{2}{*}{$I\rightarrow T$} & lin & \textbf{0.259} & 0.149 & \textbf{0.619} & 0.133 \\
		&  & nn & \textbf{0.212} & 0.147 & \textbf{0.262} & 0.144 \\ \cline{2-7} 
		& \multirow{2}{*}{$T\rightarrow I$} & lin & \textbf{0.527} & 0.147 & \textbf{0.496} & 0.137 \\
		&  & nn & \textbf{0.578} & 0.143 & \textbf{0.51} & 0.135 \\ \hline
	\end{tabular}
	\caption{Test mean nearest neighbor overlap with Euclidean-based neighbors and \textit{cosine loss}.}
	\label{tab:exp1_cosine_loss_euclidean}
\end{table}

\subsubsection{Results with Euclidean neighbors (\textit{nn} and \textit{lin} models of the paper)}

\begin{table}[H]
	\center
	\begin{tabular}{@{}lcccccc@{}}
		\toprule
		&  &  & \multicolumn{2}{c}{ResNet} & \multicolumn{2}{c}{VGG-128} \\ \midrule
		&  &  & $X,f(X)$ & $Y,f(X)$ & $X,f(X)$ & $Y,f(X)$ \\ \midrule
		\multirow{4}{*}{\rotatebox[origin=c]{90}{ImageNet}} & \multirow{2}{*}{$I\rightarrow T$} & lin & \textbf{0.671} & 0.228 & \textbf{0.695} & 0.209 \\
		&  & nn & \textbf{0.61} & 0.234 & \textbf{0.665} & 0.219 \\ \cmidrule(l){2-7} 
		& \multirow{2}{*}{$T\rightarrow I$} & lin & \textbf{0.372} & 0.233 & \textbf{0.326} & 0.218 \\
		&  & nn & \textbf{0.332} & 0.258 & \textbf{0.298} & 0.242 \\ \midrule
		\multirow{4}{*}{\rotatebox[origin=c]{90}{IAPR TC}} & \multirow{2}{*}{$I\rightarrow T$} & lin & \textbf{0.341} & 0.194 & \textbf{0.385} & 0.156 \\
		&  & nn & \textbf{0.3} & 0.203 & \textbf{0.318} & 0.17 \\ \cmidrule(l){2-7} 
		& \multirow{2}{*}{$T\rightarrow I$} & lin & \textbf{0.504} & 0.188 & \textbf{0.431} & 0.156 \\
		&  & nn & \textbf{0.421} & 0.21 & \textbf{0.363} & 0.169 \\ \midrule
		\multirow{4}{*}{\rotatebox[origin=c]{90}{Wiki}} & \multirow{2}{*}{$I\rightarrow T$} & lin & \textbf{0.245} & 0.146 & \textbf{0.235} & 0.141 \\
		&  & nn & \textbf{0.261} & 0.151 & \textbf{0.269} & 0.143 \\ \cmidrule(l){2-7} 
		& \multirow{2}{*}{$T\rightarrow I$} & lin & \textbf{0.564} & 0.149 & \textbf{0.555} & 0.135 \\
		&  & nn & \textbf{0.539} & 0.149 & \textbf{0.529} & 0.14 \\ \bottomrule
	\end{tabular}
	\caption{Test mean nearest neighbor overlap with Euclidean-based neighbors and MSE loss. Boldface indicates best performance between each $\text{\textit{mNNO}}^{10}(X,f(X))$ and  $\text{\textit{mNNO}}^{10}(Y,f(X))$ pair, which are abbreviated by $X,f(X)$ and $Y,f(X)$.}
	\label{tab:exp1_euclidean}
\end{table}

\begin{table}[H]
	\center
	\begin{tabular}{@{}cccccc@{}}
		\toprule
		&  & \multicolumn{2}{c}{ResNet} & \multicolumn{2}{c}{VGG-128} \\ \midrule
		&  & $X,f(X)$ & $Y,f(X)$ & $X,f(X)$ & $Y,f(X)$ \\ \midrule
		\multirow{2}{*}{$I\rightarrow T$} & lin & \textbf{0.57} & 0.16 & \textbf{0.644} & 0.159 \\
		& nn & \textbf{0.546} & 0.179 & \textbf{0.64} & 0.171 \\ \midrule
		\multirow{2}{*}{$T\rightarrow I$} & lin & \textbf{0.325} & 0.206 & \textbf{0.283} & 0.2 \\
		& nn & \textbf{0.283} & 0.236 & \textbf{0.259} & 0.223 \\ \bottomrule
	\end{tabular}
	\caption{Test $\text{\textit{mNNO}}$ with Euclidean-based neighbors in \textbf{ImageNet} dataset, using \textit{word2vec} word embeddings.}
	\label{tab:appendix_imagenet_word2vec}
\end{table}

\subsubsection{Results with word2vec in ImageNet (cosine-based neighbors)}

\begin{table}[H]
	\center
	\begin{tabular}{@{}cccccc@{}}
		\toprule
		&  & \multicolumn{2}{c}{ResNet} & \multicolumn{2}{c}{VGG-128} \\ \midrule
		&  & $X,f(X)$ & $Y,f(X)$ & $X,f(X)$ & $Y,f(X)$ \\ \midrule
		\multirow{2}{*}{$I\rightarrow T$} & lin & \textbf{0.61} & 0.232 & \textbf{0.674} & 0.221 \\
		& nn & \textbf{0.578} & 0.253 & \textbf{0.666} & 0.236 \\ \midrule
		\multirow{2}{*}{$T\rightarrow I$} & lin & \textbf{0.364} & 0.213 & \textbf{0.348} & 0.21 \\
		& nn & \textbf{0.356} & 0.245 & \textbf{0.331} & 0.234 \\ \bottomrule
	\end{tabular}
	\caption{Test $\text{\textit{mNNO}}$ using cosine-based neighbors in \textbf{ImageNet}, using \textit{word2vec} word embeddings.}
	\label{tab:exp1_imagenet_word2vec}
\end{table}

\subsection{Experiment 2} 

\begin{table}[H]
	\center
	\begin{tabular}{@{}lcccccc@{}}
		\toprule
		& \multicolumn{2}{c}{WS-353} & \multicolumn{2}{c}{Men} & \multicolumn{2}{c}{SemSim} \\ \cmidrule(l){1-7} 
		& Cos & Eucl & Cos & Eucl & Cos & Eucl \\ \midrule
		$f_{\text{nn}}$(word2vec) & 0.665 & 0.636 & 0.782 & 0.781 & 0.729 & 0.719 \\
		$f_{\text{lin}}$(word2vec) & 0.67 & 0.527 & 0.785 & 0.696 & 0.737 & 0.616 \\
		word2vec & 0.669 & 0.533 & 0.787 & 0.701 & 0.742 & 0.62 \\ \midrule
		$f_{\text{nn}}$(VGG-128) & 0.44 & 0.433 & 0.588 & 0.585 & 0.521 & 0.513 \\
		$f_{\text{lin}}$(VGG-128) & 0.445 & 0.301 & 0.593 & 0.496 & 0.531 & 0.344 \\
		VGG-128 & 0.448 & 0.307 & 0.593 & 0.496 & 0.534 & 0.344 \\ \midrule
		&  &  &  &  &  &  \\ \midrule
		& \multicolumn{2}{c}{VisSim} & \multicolumn{2}{c}{SimLex} & \multicolumn{2}{c}{SimVerb} \\ \cmidrule(l){1-7} 
		& Cos & Eucl & Cos & Eucl & Cos & Eucl \\ \midrule
		$f_{\text{nn}}$(word2vec) & 0.566 & 0.567 & 0.419 & 0.379 & 0.309 & 0.232 \\
		$f_{\text{lin}}$(word2vec) & 0.572 & 0.507 & 0.429 & 0.275 & 0.328 & 0.174 \\
		word2vec & 0.576 & 0.51 & 0.435 & 0.279 & 0.308 & 0.15 \\ \midrule
		$f_{\text{nn}}$(VGG-128) & 0.551 & 0.547 & 0.404 & 0.399 & 0.231 & 0.235 \\
		$f_{\text{lin}}$(VGG-128) & 0.56 & 0.404 & 0.406 & 0.335 & 0.23 & 0.316 \\
		VGG-128 & 0.56 & 0.403 & 0.406 & 0.335 & 0.235 & 0.329 \\ \bottomrule
	\end{tabular}
	\caption{Spearman correlations between human ratings and similarities (cosine or Euclidean) predicted from the embeddings, using \textit{word2vec} and \textit{VGG-128} embeddings.}
	\label{tab:exp2_word2vec_vgg128}
\end{table}

\end{document}